# Deep Self-representative Concept Factorization Network for Representation Learning

Yan Zhang[1]  Zhao Zhang[1,2,*]  Zheng Zhang[3]  Mingbo Zhao[4]  Li Zhang[1]  Zhengjun Zha[5]  Meng Wang[2]

**Abstract** — In this paper, we investigate the unsupervised deep representation learning issue and technically propose a novel framework called *Deep Self-representative Concept Factorization Network* (DSCF-Net), for clustering deep features. To improve the representation and clustering abilities, DSCF-Net explicitly considers discovering hidden deep semantic features, enhancing the robustness properties of the deep factorization to noise and preserving the local manifold structures of deep features. Specifically, DSCF-Net seamlessly integrates the robust deep concept factorization, deep self-expressive representation and adaptive locality preserving feature learning into a unified framework. To discover hidden deep representations, DSCF-Net designs a hierarchical factorization architecture using multiple layers of linear transformations, where the hierarchical representation is performed by formulating the problem as optimizing the basis concepts in each layer to improve the representation indirectly. DSCF-Net also improves the robustness by subspace recovery for sparse error correction firstly and then performs the deep factorization in the recovered visual subspace. To obtain locality-preserving representations, we also present an adaptive deep self-representative weighting strategy by using the coefficient matrix as the adaptive reconstruction weights to keep the locality of representations. Extensive comparison results with several other related models show that DSCF-Net delivers state-of-the-art performance on several public databases.

## I. Introduction

Representation learning from high-dimensional complex data is always an important and fundamental problem in the fields of pattern recognition and data mining [40-50]. To represent data, lots of feasible and effective approaches can be used, of which Matrix Factorization (MF) based models have been proven to be effective for low-dimensional feature extraction and clustering [24-32][36-39]. Nonnegative Matrix Factorization (NMF) [1] and Concept Factorization (CF) [2] are two most classical nonnegative MF methods. Given a nonnegative data matrix $X$, both NMF and CF aim to decompose it into the product of two or three nonnegative factors by minimizing the reconstruction error. To be more specific, one factor contains the basis vectors capturing the higher-level features of data and each sample can be reconstructed by a linear combination of the bases. The other factor corresponds to the new low-dimensional representation. Since the nonnegative constraints are applied in NMF and CF, they can obtain local parts-based representations [1]. It is noteworthy that those distinguishing features may be precisely reflected by the key parts (such as eyebrows and ears in face images; directions of textures in texture images and the angular angles in graphs) in reality. As such, the nonnegative constraint can play an essential role in feature representation.

For obtaining the locality preserving feature representations, many MF based methods usually adopt the graph regularization strategy, such as Graph Regularized NMF (GNMF) [3], Locally Consistent CF (LCCF) [4], Self-Representative Manifold Concept Factorization (SRMCF) [5], and some dual-graph regularized methods, e.g., Dual Regularization NMF (DNMF) [6] and Dual-graph regularized CF (GCF) [7]. Specifically, GNMF and LCCF apply the graph Laplacian to smooth the representation and encode the geometrical information of the data space, which allows extracting the new representation with respect to the intrinsic manifold structures. SRMCF constructs the affinity matrix by assigning adaptive neighbors to each sample based on the local distance of learned new representation of the original data with itself as a dictionary [5]. Different from GNMF, LCCF and SRMCF, both DNMF and GCF not only preserve the geometric structures of data manifold but also the feature manifold jointly using the dual-graph regularization. In addition to the above graph regularization strategy, another way to retain the locality is by the local coordinate coding. Local Coordinate CF (LCF) [8], Graph-Regularized LCF (GRLCF) [9] and Graph-Regularized CF with Local Coordinate (LGCF) [10] are several classical methods. The local coordinate coding can enable each sample to be represented of a linear combination with only a few nearby basis concepts so that the locality and sparsity can be discovered simultaneously [8-10].

Although the above-mentioned methods have obtained enhanced representation performance, they still suffer from some drawbacks. (1) Most existing MF based methods aim at factorizing the data in the original visual space that usually contains noise and corruptions, which may directly result in the degraded performance; (2) To preserve the locality of the learned new representation, graph regularization based algorithms, including GNMF, DNMF, LCCF, GRLCF, GCF and LGCF, need to look for the neighbors of each sample by the $k$-neighborhood or $\varepsilon$-neighborhood. However, it still remains a tricky problem to estimate an optimal $k$ or $\varepsilon$ in practice. Moreover, they often pre-compute the graph weights by a separable process before learning the new representations, but such operation cannot ensure the pre-obtained weights as joint-optimal for low-dimensional representation learning. Although recent SRMCF employs the self-representation of original data to construct the affinity matrix to preserve the locality of coefficients, but it also involves an extra graph regularization based on the original data to encode the manifold structures of new representations, so it will clearly suffers from the same tricky issues as GNMF, DNMF, LCCF, GRLCF, GCF and LGCF; (3) The last and most important point is that aforementioned methods are essential single-layer models (i.e., 1-layer), so they can only obtain shallow features. That is, they cannot discover deep semantic features from data, since shallow models directly map visual samples into a latent subspace, while such an operation will implicitly neglect and lose certain important hidden information.

[1] School of Computer Science and Technology & Provincial Key Laboratory for Computer Information Processing Technology, Soochow University, Suzhou, China. Emails: zhangyan0712suda@gmail.com, zhangliml@suda.edu.cn.
[2] Key Laboratory of Knowledge Engineering with Big Data (Ministry of Education) & School of Computer Science and Information Engineering (School of Artificial Intelligence), Hefei University of Technology, Hefei, China. Emails: cszzhang@gmail.com, eric.mengwang@gmail.com.
[3] Department of Computer Science, Harbin Institute of Technology (Shenzhen), Shenzhen, China. Email: darrenzz219@gmail.com.
[4] Department of Electronic Engineering, City University of Hong Kong, Tat Chee Avenue, Kowloon, Hong Kong. Email: mzhao4-c@my.cityu.edu.hk.
[5] School of Information Science and Technology, University of Science and Technology of China, Hefei, China. Email: zhazj@ustc.edu.cn.
* indicates the corresponding author.

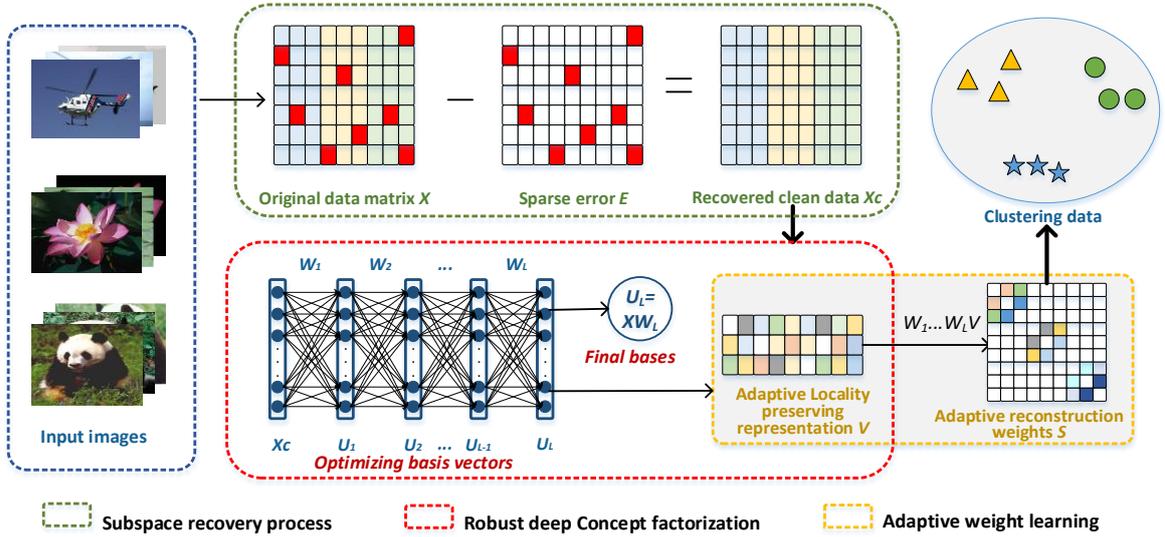

Figure 1: The hierarchical architecture of our proposed DSCF-Net framework.

In more recent years, researchers have also investigated the topics of multilayer matrix factorization. One most commonly-used approach of extending the shallow model to deep model is to decompose the observation data matrix iteratively in a number of layers by a cascade connection of $L$ mixing subsystems ($L$ is the number of layers in the deep networks) [11-14]. Multi-layer NMF (MNMF) [11], Multilayer CF (MCF) [12], Spectral Unmixing using Multilayer NMF (MLNMF) [13] and Graph Regularized Multilayer CF (GMCF) [14] are some representative models in this category. These multilayer MF models directly take the outputted new representation of a previous layer as the input of the next layer, but this strategy may be ineffective and even unreasonable in reality, since we cannot ensure that the output of previous layer is an optimal representation of the original data and feeding the output of the previous layer directly to next layer may mislead and degrade the representation learning process of the subsequent layers. Moreover, since the learned representations in the first layer may be inaccurate and may lose important hidden information, the resulted reconstruction errors may be getting larger and larger with the increasing of layers. Besides, by simply transferring the new representation to the next layer, they still need to initialize the basis vectors randomly in each layer, which is also unreasonable empirically. To address these issues, Weakly-supervised Deep MF (WDMF) [15] and Deep Semi-NMF (DSNMF) [16] have provided another way to construct the deep NMF models. Specifically, they define the deep network models to discover hidden deep feature information by multiple layers of linear transformations and update the basis concepts/new representations in each layer. In this way, WDMF and DSNMF can obtain the hidden deep representation explicitly. But noting that WDMF and DSNMF still cannot encode the local geometry structure of the new representations by self-expression in each layer explicitly in an adaptive manner. They also factorize data in the original input space that usually has various noise and corruptions that may decrease the performance. We argue that the descriptive abilities and quality of learned feature representations from the first layer will be critical for the subsequent layers.

In this paper, we mainly propose certain effective strategies to overcome the drawbacks of existing MF models and obtain more powerful deep representations for unsupervised clustering. The main contributions of this work are summarized as

(1) A novel Deep Self-representative Concept Factorization network, termed DSCF-Net, is technically proposed for deep feature learning and clustering. DSCF-Net explicitly considers improving the feature representation by mining deep semantic features hidden in data, enhancing the robustness properties of the learning system to noise and preserving the local manifold structures of deep features in an adaptive manner. Fig.1 illustrates the flowchart of our DSCF-Net. We see that DSCF-Net seamlessly integrates the robust subspace recovery, robust deep concept factorization, self-expressive representation learning and adaptive locality preservation into a united framework.

(2) To deliver a better higher-level deep feature representation and well handle the semantic gap, on one hand DSCF-Net designs a hierarchical factorization architecture using the multiple layers of linear transformations to obtain the latent representation by a progressive way. Such an operation can automatically learn the intermediate hidden representations and update the intermediate basis vectors in each layer. Because the basis vectors capture the higher-level features of input data and each sample is reconstructed using a linear combination of the bases, we argue that optimizing the basis vectors to improve the representation indirectly may be more important than optimizing the new representations in each layer. Meanwhile, DSCF-Net is modeled as the formulation of learning one final representation matrix and $L$ updated sets of basis vectors. On the other hand, our DSCF-Net learns the latent deep representation in a recovered clean subspace by leveraging the geometrical, visual and semantic information jointly. Due to the fact that the used subspace recovery process can remove noise and outliers explicitly from original data, both the robustness properties and descriptive power of the learned representation in the first layer and subsequent layers can be potentially enhanced.

(3) To obtain the locality-preserving higher-level representations, DSCF-Net introduces the adaptive self-representative weighting strategy. Specifically, our DSCF-Net explicitly uses the coefficient matrix as the adaptive reconstruction weights to preserve local information of new representation in each layer.

We outline the paper as follows. Section II briefly reviews the related works. Our DSCF-Net is presented in Section III. In Section IV, we mainly show the optimization procedures of our DSCF-Net. The comparison of the architectures of exiting single-layer and multilayer CF frameworks are discussed in Section V. Section VI describes the simulation settings and results. Finally, the paper is concluded in Section VII.

## II. RELATED WORK

In this section, we review several single-layer and multilayer methods relevant to our proposed DSCF-Net framework.

### A. Single-layer CF and SRMCF

**Concept factorization [2].** We first briefly introduce the CF model. For a given data matrix $X = [x_1, x_2, ..., x_N] \in \mathbb{R}^{D \times N}$, where $x_i, i \in 1, 2, ..., N$ is a sample vector, $N$ is the number of samples and $D$ is the original dimensionality. Denote by $U \in \mathbb{R}^{D \times r}$ and $V \in \mathbb{R}^{r \times N}$ two nonnegative matrices whose product $UV \in \mathbb{R}^{D \times N}$ is the approximation to $X$, where the rank $r$ is a constant, by representing each basis by using a nonnegative linear combination of $x_i$, i.e., $\sum_{i=1}^{N} w_{ij} x_i$ with $w_{ij} \geq 0$, CF aims at solving the following minimization problem:

$$O = \|X - XWV\|_F^2, s.t. W, V \geq 0, \quad (1)$$

where $W = [w_{ij}] \in \mathbb{R}^{N \times r}$, i.e., $XW$ approximates the bases, and $V$ is the learnt new representation of $X$ for clustering.

**Self-representative manifold CF (SRMCF) [5].** SRMCF improves CF by integrating the adaptive neighbor structure and manifold regularizers into a unified model. It is noteworthy that CF can be considered as an improved self-representation method with a learning based dictionary to reveal the global structure of data, and the coefficients of CF carry plentiful semantic meanings. By rewriting the CF model, one can have

$$X \approx XR, where\ R = WV, \quad (2)$$

where $R = WV$ is regarded as the coefficient matrix based on the dictionary using the original data.

To formulate the model, SRMCF involves two similarity matrices $S$ and $A$ to preserve the locality of new representation $V$ and coefficient matrix $WV$, respectively. Let $L_S$ and $L_A$ denote Laplacian matrices, i.e., $L_S=D_S-S$, $L_A=D_A-A$, where $D_S$ and $D_A$ are two diagonal matrices whose entries are column sums of the similarity matrices $S$ and $A$ respectively. $S$ is defined based on the binary-weighting in the original data space as

$$S_{ij} = \begin{cases} 1 & if\ x_i \in N_k(x_j)\ or\ x_j \in N_k(x_i) \\ 0 & otherwise \end{cases},$$

where $N_k(x_i)$ is the $k$-nearest neighbor set of $x_i$. $A$ can be obtained adaptively based on the coefficient matrix $WV$ by solving the following problem:

$$\min_{A_i^T \mathbf{1}=1,\ 0 \leq A_i \leq 1} \sum_{j=1}^{N} \|(WV)_i - (WV)_j\|_2^2 A_{ij} + \gamma A_{ij}^2,$$

where $\gamma$ is a positive parameter and $\mathbf{1}$ is a vector of all ones. Finally, the objective function of SRMCF is formulated as

$$\min_{W,V,A} \|X - XWV\|_F^2 + \lambda_1 tr(WVL_A VW) + \lambda_2 tr(VL_S V^T) + \lambda_3 \|A\|_F^2, \\ s.t.\ W, V \geq 0, \forall_i\ A_i^T e = 1,\ 0 \leq A_i \leq 1 \quad (3)$$

where $\lambda_1$, $\lambda_2$, and $\lambda_3$ are nonnegative tunable parameters. Note that although SRMCF defines the affinity matrix $A$ in an adaptive manner, it still uses the traditional weighting method to define $S$, so it also faces the difficult issue of identifying $k$ in reality. Note that DSCF-Net avoids this tricky issue by directly using the coefficient matrix as the adaptive reconstruction weight matrix for encoding the locality of $V$.

### B. Weakly-supervised Deep Matrix Factorization (WDMF)

We briefly review the deep architecture of WDMF. Assume that the hierarchical structure has $L$ layers, WDMF factorizes the observed image tagging matrix $F$ into $L+1$ factor matrices, i.e., $U, V_L, ..., V_1$, and the output of first layer is transformed from visual space, i.e., $V_1=W_1X$. Specifically, WDMF applies a deep network to discover the hidden representations as

$$\begin{aligned} F &\leftarrow UV_L \\ V_L &= W_L V_{L-1} \\ &\vdots \\ V_2 &= W_2 V_1 \\ V_1 &= W_1 X \end{aligned}, \quad (4)$$

where $W_l$ ($l=1,2,..., L$) is the transformation matrix of the $l$-th layer, $U$ is the latent tag feature matrix in the subspace and $V_l$ is the implicit representation matrix of images in the $l$-th layer. That is, the problem of WDMF learns one factor $U$ containing the basis vectors and $L$ representation matrices $V_L, ..., V_1$. The unified objective function of WDMF is defined as

$$\min_{V,W_L,\ldots,W_1} \frac{1}{2}\|F-UV\|_F^2 + \frac{\alpha}{2}tr\left(V_{L-1}GV_{L-1}^T\right) + \frac{\beta}{2}tr\left(U^T PU\right)$$
$$+\frac{\mu}{2}tr\left(VMV^T\right) + \frac{\lambda_1}{2}\left(\|U\|_F^2 + \sum_{l=1}^L W_l\right) + \frac{\lambda_2}{2}\|W_1\|_{2,1} \quad , \quad (5)$$
$$s.t. \ W_l^T W_l = I$$

where $V=W_L W_{L-1} \ldots W_1 X$, $G=D_B-B$ and $P=(H-I)(H-I)$ are positive semi-definite Laplacian matrices. $B_{i,j}$ measures the semantic relevance between the $i$-th image and the $j$-th image, which is defined by using the cosine similarity based on the tagging vectors and $D_B$ a diagonal matrix over $B$. $H$ is defined as: $H_{i,j} = g_{i,j}$ if $x_j \in N_k(x_i)$ and $H_{i,j}=0$ otherwise. The local discriminative structure can be well retained, which can also solve the overfitting caused by noisy tagging information. $\alpha$, $\beta$ and $\mu$ are positive trade-off factors, $\lambda_1$ and $\lambda_2$ are regularization factors.

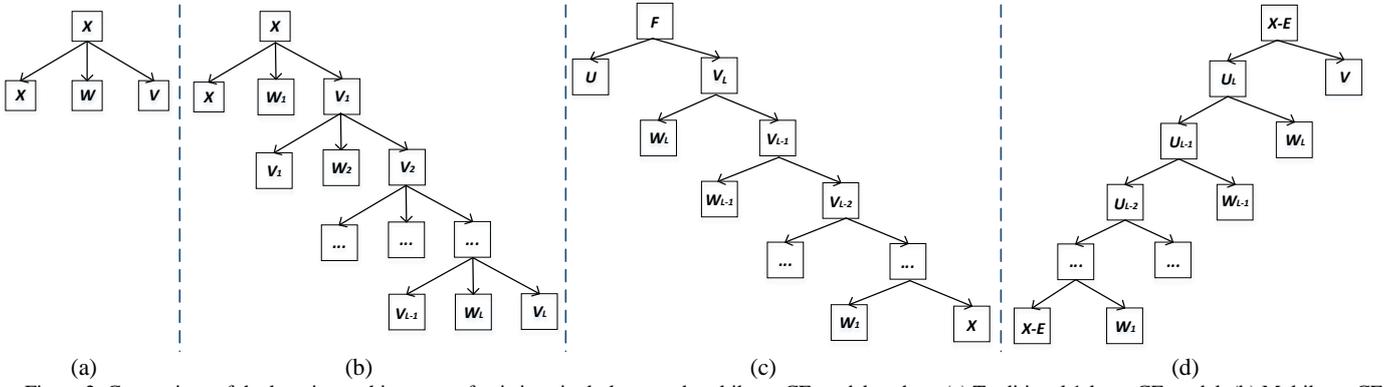

(a) (b) (c) (d)

Figure 2: Comparison of the learning architectures of existing single-layer and multilayer CF models, where (a) Traditional 1-layer CF model; (b) Multilayer CF model (e.g., MNMF, MCF, MLNMF and GMCF); (c) WDMF model; (d) Our DSCF-Net model.

## III. DEEP SELF-REPRESENTATIVE CONCEPT FACTORIZATION NETWORK (DSCF-NET)

We introduce the formulation of our DSCF-Net in this section. Given data matrix $X=[x_1,x_2,\ldots,x_N]\in\mathbb{R}^{D\times N}$, we design a deep hierarchical structure that has $L$ layers. Since the basis vectors capture the higher-level features of data and each sample is reconstructed by a linear combination of the bases, DSCF-Net aims to find $L+1$ nonnegative matrices, and is formulated as the problem of learning one representation matrix $V$ and $L$ sets of basis vectors, i.e., $U_1, U_2, \ldots, U_L$.

Since real-world original data $X$ usually have various noise and errors, DSCF-Net also considers improving the robustness properties to noise by using subspace recovery. Specifically, DSCF-Net performs the hierarchical factorization over recovered clean data $X-E$ rather than the original data $X$ so that the representation ability can be enhanced, where $E$ is the sparse error by L2,1-norm regularization, i.e., $\|E\|_{2,1}$. As a result, the factorization process of our DSCF-Net is obtained as

$$(X-E) \leftarrow U_L V$$
$$U_L = U_{L-1} W_L$$
$$\vdots \quad , \quad (6)$$
$$U_2 = U_1 W_2$$
$$U_1 = (X-E) W_1$$

where $U_l$ ($l=1, 2,\ldots, L$) is the set of basis vectors of the $l$-th layer in the recovered clean visual subspace and $W_l$ ($l=1, 2,\ldots, L$) is the intermediate matrix for updating the basis vectors. Finally, the product $U_L V$ that is equivalent to $(X-E)W_1\ldots W_L V$ is the reconstruction of the original data. In other words, the total reconstruction error can be defined as

$$J_1 = \|(X-E)-(X-E)W_0\ldots W_{L-1}W_L V\|_F^2, \quad (7)$$

where $(X-E)W_0\ldots W_{L-1}W_L$ is called deep basis vectors, $V$ is the new deep representation of the input data and $W_0$ is fixed to be an identity matrix. Following SRMCF [5], the reconstruction can be rewritten as $(X-E)=(X-E)R$, where $R=W_0\ldots W_{L-1}W_L V$ can be similarly regarded as the meaningful coefficient matrix based on the clean dictionary of recovered data matrix $X-E$, and the coefficients can be used to characterize the locality and similarity between samples or features. Note that the coefficient matrix $UV$ of SRMCF is directly based on the dictionary of the input $X$ that usually contains noise, so the resulted coefficient matrix of our DSCF-Net will be potentially more accurate than that of SRMCF in practice. Moreover, we adopt a different weighting strategy as SRMCF. That is, we directly use the coefficient matrix $R=W_0\ldots W_{L-1}W_L V$ in each layer as the adaptive reconstruction weights to encode and preserve the locality of learned new representation $V$, which is clearly different from SRMCF that involves an extra graph regularization on the original data to retain the locality of $V$. Then the locality preserving constraint of DSCF-Net is defined as

$$J_2 = \|V-V(W_0\ldots W_{L-1}W_L V)\|_F^2 + \|W_L V\|_F^2. \quad (8)$$

Based on the reconstruction error $J_1$ and locality preserving constraint $J_2$, the objective function of DSCF-Net is defined as

$$\min_{W_1,\ldots,W_L,V,E} J_1 + \alpha J_2 + \gamma \|E\|_{2,1}$$
$$= \|(X-E)-(X-E)W_0\ldots W_{L-1}W_L V\|_F^2$$
$$+\alpha\left(\|V-V(W_0\ldots W_{L-1}W_L V)\|_F^2 + \|W_L V\|_F^2\right) + \gamma\|E\|_{2,1} \quad , \quad (9)$$
$$s.t. \ \forall_{l\in\{1,2,\ldots,L\}} W_l \geq 0, V \geq 0$$

where $\alpha,\gamma \geq 0$ are trade-off parameters, the L2,1-norm $\|E\|_{2,1} = \sum_{j=1}^N \sqrt{\sum_{i=1}^n |E_{i,j}|^2}$ can make the error term $E$ column sparse. To facilitate the optimization, we include an auxiliary variable $S$

to relax the locality preserving constraint. The relaxed optimization problem can be written as

$$\min_{W_1,...,W_L,V,E,S} \|(X-E)-(X-E)W_0...W_{L-1}W_LV\|_F^2$$
$$+ \alpha\left(\|S-W_1...W_{L-1}W_LV\|_F^2 + \|W_LV\|_F^2\right)$$
$$+ \beta\|V-VS\|_F^2 + \gamma\|E\|_{2,1} \quad (10)$$
$$s.t. \quad \forall_{l\in\{1,2,...,L\}} W_l \geq 0, V \geq 0$$

where $\beta \geq 0$ is also a trade-off parameter. Next, we detail the optimization procedures of our DSCF-Net.

## IV. OPTIMIZATION

From the objective function of DSCF-Net, we can find that the involved several variables, i.e., $W_l$ ($1 \leq l \leq L$), $V$, $S$ and $E$, depend on each other, so they cannot be solved directly. Following the common procedures, we propose an iterative optimization strategy by using the Multiplicative Update Rules (MUR) method for local optimal solutions. Specifically, we solve the problem by updating the variables alternately and compute one of the variables each time by fixing others. The detailed optimization procedures of DSCF-Net are shown as follows:

**1) Fix others, update the factors $W_l$ and $V$:**

When the other variables are fixed, we can update the matrices $W_l$ and $V$ by solving the objective function. For the $l$-th layer, $W_1,...,W_{l-1}$ are all constants, and we define $\Pi_{l-1}=W_0...W_{l-1}$. By using $X_c$ to denote the recovered clean data, i.e., $X-E$, the problem associated with $W_l$ and $V$ can be defined as

$$\min_{W_l,V} \|X_c - X_c\Pi_{l-1}W_lV\|_F^2 + \alpha\left(\|S-\Pi_{l-1}W_lV\|_F^2 + \|W_lV\|_F^2\right)$$
$$+ \beta tr(V^TQV), \quad s.t. \quad W_l \geq 0, V \geq 0 \quad (11)$$

where $Q=(I-S)(I-S)^T$ and $I$ is an identity matrix. Let $\psi_{ik}$ and $\phi_{ik}$ be the Lagrange multipliers for the constraints $(W_l)_{ik} \geq 0$ and $v_{ik} \geq 0$, and $\Psi=[\psi_{ik}]$, $\Phi=[\phi_{ik}]$, the Lagrange function of the above problem can be constructed as

$$\min_{W_l,V} \|X_c - X_c\Pi_{l-1}W_lV\|_F^2 + \alpha\left(\|S-\Pi_{l-1}W_lV\|_F^2 + \|W_lV\|_F^2\right)$$
$$+ \beta tr(V^TQV) + tr(\Psi W_l^T) + tr(\Phi V^T) \quad (12)$$

For ease of representation, we use notation $O_1$ to denote the objective function of the above problem. Then, the variables $W_l$ and $V$ can be alternately updated by fixing other variables. The derivatives of $O_1$ w.r.t. $W_m$ and $V$ are computed as follows:

$$\partial O_1/\partial W_l = 2\left(\Pi_{l-1}^T K_c WVV^T - \Pi_{l-1}^T K_c V^T\right)$$
$$+ 2\alpha\left(-\Pi_{l-1}SV^T + \Pi_{l-1}^T\Pi_{l-1}WVV^T\right) + \Psi \quad (13)$$

$$\partial O_1/\partial V = 2\left(W^T\Pi_{l-1}^t K_c\Pi_{l-1}WV - W^T\Pi_{l-1}^T K_c\right)$$
$$+ 2\alpha\left(W^T\Pi_{l-1}^t\Pi_{l-1}WV + W^TWV - W^T\Pi_{l-1}^T S\right) \quad (14)$$
$$+ 2\beta VQ^T + \Phi$$

where $K_c = X_c^T X_c$. By using the KKT conditions $\psi_{ik}(w_l)_{ik}=0$ and $\phi_{ik}v_{ik}=0$, we can obtain the updating rules as

**Algorithm 1: Optimization procedures of DSCF-Net**
**Inputs:** Training data $X$, constant $r$ and parameters $\alpha, \beta, \gamma$.
**Initialization:** Initialize the variables $W$, and $V$ as random matrices; initialize $S$ and $E$ as zero matrices; initialize $D$ to be an identity matrix; $\varepsilon=10^{-3}$; $t=0$.
*For fixed number $l$ of layers:*
**While not converged do**
1. Update the matrix factors $W_l$ and $V$ by Eqs.(15-16);
2. Update the auxiliary matrix $S$ by Eq.(18);
3. Update the sparse error $E$ by Eq.(21) and update the entries of the diagonal matrix $D$ by $d_{ii}=1/(2\|e_i\|_2)$;
4. Check for convergence: if $\|V^{t+1}-V^t\|_F \leq \varepsilon$, stop; else $t=t+1$.
**End while**
**Output:** The learned deep new representations $V^*$.

$$(w_l)_{ik} \leftarrow (w_l)_{ik}\frac{\left(\Pi_{l-1}^T K_c V^T + \alpha\Pi_{l-1}^T SV^T\right)_{ik}}{\left(\Pi_{l-1}^T K_c\Pi_{l-1}^T WVV^T + \alpha\Pi_{l-1}^T\Pi_{l-1}WVV^T + \alpha WVV^T\right)_{ik}}, \quad (15)$$

$$v_{ik} \leftarrow v_{ik}\frac{\left(W^T\Pi_{l-1}^T K_c + \alpha W^T\Pi_{l-1}^T S\right)_{ik}}{\left(W^T\Pi_{l-1}^T K_c\Pi_{l-1}^T WV + \alpha W^T\Pi_{l-1}^T\Pi_{l-1}WV + \alpha W^T WV + \beta VQ^T\right)_{ik}}. \quad (16)$$

**2) Fix others, update the auxiliary variable $S$:**
With obtained $W_l$ and $V$, we can use them to update the auxiliary variable $S$ by solving the following problem:

$$\min_S O_2 = \alpha\|S-\Pi_l V\|_F^2 + \beta\|V-VS\|_F^2$$
$$= \alpha tr\left((S-\Pi_l V)(S-\Pi_l V)^T\right) + \beta tr\left((V-VS)(V-VS)^T\right) \quad (17)$$

where $\Pi_l = W_1...W_{l-1}W_l$. Then, the variable $S$ can be obtained by setting the derivative of $O_2$ w.r.t $S$ to zero:

$$\partial O_2/\partial S = \alpha(S-\Pi_l V) + \beta(V^T VS - V^TV) = 0$$
$$\Rightarrow S = (\alpha I + \beta V^TV)^{-1}(\alpha\Pi_l V + \beta V^TV) \quad (18)$$

**3) Fix others, recover the sparse error $E$:**
After calculating $W_l$, $V$ and $S$, we can easily update the sparse errors $E$ by solving the following reduced formulation:

$$\min_E O_3 = \|(X-E)-(X-E)W_0...W_{l-1}W_lV\|_F^2 + \gamma\|E\|_{2,1}. \quad (19)$$

By the properties of $L_{2,1}$-norm, we have $\|E\|_{2,1}=2tr(E^TDE)$, where $D$ is a diagonal matrix with $d_{ii}=1/(2\|e_i\|_2)$ being its entries, and $e_i$ is the $i$-th column vector of $E$. If each $e_i \neq 0$, the above formulation can be approximated as

$$\min_E O_3 = \|(X-E)(I-W_0...W_{l-1}W_lV)\|_F^2 + \gamma tr(E^TDE). \quad (20)$$

Let $\Theta_l = I - W_0...W_{l-1}W_lV$, then we can update $E$ by computing the derivative of the above problem $O_3$ w.r.t. $E$ as

$$\partial O_3/\partial E = 2(E-X)\Theta_l\Theta_l^T + 2\gamma ED = 0$$
$$\Rightarrow E = X\Theta_l\Theta_l^T\left(\Theta_l\Theta_l^T + \gamma D\right)^{-1} \quad (21)$$

For complete presentation of our DSCF-Net, we summarize the optimization procedures of our DSCF-Net in Algorithm 1, where the diagonal matrix $D$ is initialized as an identity matrix.

DSCF-Net mainly optimizes the basis vectors to improve the representation $V$ that is the major variable. To ensure the representation $V$ to converge, the stopping condition is simply set to $\|V^{t+1}-V^t\|_F^2 \leq \varepsilon$ in each layer, which measures the difference between sequential representation matrices $V$s and it can make sure that the representation result will not change drastically.

## V. Discussion and Some Remrks

In this section, we mainly compare the architectures of exiting single-layer and multilayer CF frameworks in Fig.2. As shown in Fig.2(a), the single-layer CF model obtains the basis vectors and new representation directly from given data by 1-layer, so it fails to discover the deep hidden semantic and structure information. As shown in Fig.2(b), those traditional multilayer CF models, e.g., MNMF, MCF, MLNMF and GMCF, directly use the output of previous layer (i.e., intermediate representation $V$) as the input of subsequent layer, without properly considering the optimization of new representation and basis vectors in each layer. Besides, since one cannot ensure the intermediate representation $V$ from the previous layer to be a good representation for subsequent layer, which may cause the degraded results. Different from traditional single-layer and multilayer CF models, WDMF and DSCF-Net explicitly consider optimizing the new representation in each layer, as can be seen from Figs.2(c) and (d). But it should be noted that our DSCF-Net differs from WDMF in several aspects. First, the strategies of optimizing the new representation in each layer are different. Specifically, WDMF aims at fixing the basis vectors and optimizes the representation $V$ directly in each layer, while our DSCF-Net aims at optimizing the basis vectors to improve the new representation $V$ indirectly in each layer. The major benefit of this strategy used in DSCF-Net is that the basis vectors capture the higher-level features of samples, so we believe that the procedure of reconstructing given sample by a linear combination of the bases will be more accurate if we can obtain a set of optimal basis vectors, which will also be verified by simulations. Second, their learning tasks are different. WDMF mainly focuses on the social image understanding tasks, i.e., tag refinement, tag assignment and image retrieval, and the initial input of WDMF is the tagging matrix $F$ rather than image data. While DSCF-Net mainly extracts new feature representations from the original images and the input is image data. Third, the locality preserving strategies are different. From the definition of $H$, it is clear that WDMF still suffers from the tricky issue of selecting $k$ in reality. In addition, $B$ and $H$ are pre-calculated based on the image tagging matrix and samples respectively, it is not guaranteed to obtain optimal results of subsequent representation learning. While the locality of the new representation in DSCF-Net is not pre-calculated, since it is jointly optimized in our model. Furthermore, the locality the new representation is preserved adaptively in our model.

## VI. Experimental Results and Analysis

In this section, we mainly perform simulations to examine the data representation and clustering performance of our DSCF-Net. The results of our DSCF-Net are mainly compared with several traditional single-layer matrix factorization techniques (i.e., GNMF [3], DNMF [6], LCCF [4], LCF [8]) and several deep factorization models (i.e., MNMF [11], MLNMF [13], MCF [12] and GMCF [14]). For fair comparison, the parameters of each method are carefully chosen from the candidate set, and the averaged results are reported.

In our study, three public image databases are involved, i.e., ETH80 object database [22], COIL100 object database [23] and MIT CBCL face database [21]. Following the common procedures, all the face and object images are resized into 32×32 pixels, i.e., each image is represented by a 1024-D vector. We perform all the experiments on a PC with Intel Core i5-4590 CPU @ 3.30 GHz 3.30 GHz 8G.

### A. Visual Image Analysis by Visualization

We firstly compare the locality representation power by visualizing the reconstruction weight matrix $S$ of our DSCF-Net, the binary weights used in DNMF and GNMF, and the Cosine similarity weights used in LCCF and GMCF. COIL100 object database is used and we randomly choose 200 images from first five classes to construct the adjacency graphs. The number of the nearest neighbors is set to 7 [20] for other evaluated weighting method for fair comparison. The visualization results are shown in Fig.3, where we show the adaptive weights obtained by DSCF-Net in the first and third layers. We can easily find that: 1) the constructed weight matrices by different weighting ways have approximate block-diagonal structures; 2) more noisy or wrong inter-class connections are produced in the Binary weights and Cosine Similarity weights than ours; 3) the structures of our adaptive weights in the third layer is better than the first layer, which means that our deep model can potentially improve the representation and locality.

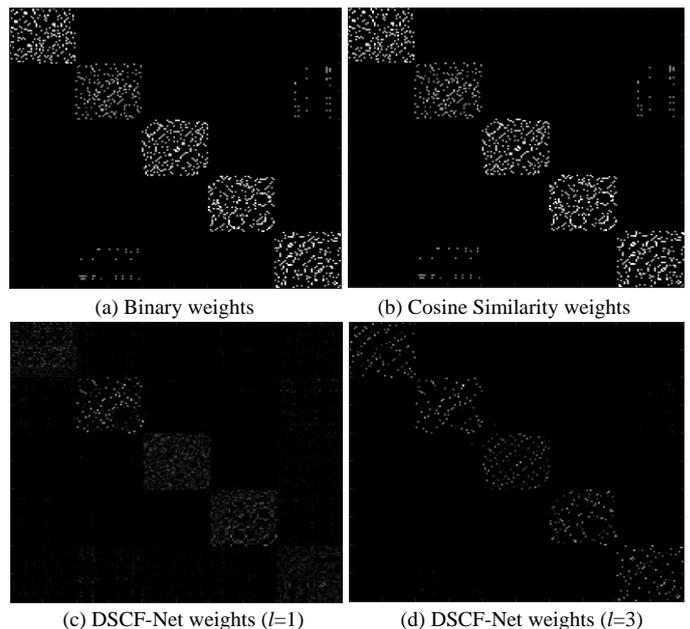

(a) Binary weights     (b) Cosine Similarity weights

(c) DSCF-Net weights ($l$=1)     (d) DSCF-Net weights ($l$=3)

*Figure 3. Visualization comparison of constructed weights.*

### B. Quantitative Clustering Evaluation

We evaluate the clustering performance of each algorithm. For quantitative evaluation, we employ two widely used clustering evaluation metrics, i.e., Accuracy (AC) and F-score [17-18].

Table 1. Clustering performance of each evaluated method based on three public image databases.

| Methods | Clustering Accuracy | | | | | | | | | | | | | | |
|---|---|---|---|---|---|---|---|---|---|---|---|---|---|---|---|
| | ETH80 object database | | | | | COIL100 object database | | | | | MIT CBCL face database | | | | |
| | K=2 | K=3 | K=4 | K=5 | K=6 | K=2 | K=3 | K=4 | K=5 | K=6 | K=2 | K=3 | K=4 | K=5 | K=6 |
| GNMF | 88.50 | 67.98 | 56.83 | 49.20 | 37.51 | 86.67 | 83.60 | 68.14 | 68.00 | 60.38 | 81.84 | 51.60 | 48.60 | 38.19 | 35.35 |
| DNMF | 87.59 | 66.91 | 56.91 | 47.74 | 37.62 | 85.38 | 84.25 | 70.39 | 67.67 | 62.35 | 82.04 | 51.91 | 46.71 | 38.86 | 34.42 |
| LCCF | 91.20 | 63.98 | 61.97 | 63.00 | 54.79 | 75.82 | 73.57 | 64.04 | 73.64 | 68.69 | 74.75 | **73.43** | 65.93 | 57.04 | 55.32 |
| LCF | 91.22 | 62.95 | 50.48 | 42.50 | 32.17 | 91.14 | 76.60 | 67.58 | 69.19 | 64.53 | 68.49 | 45.52 | 44.41 | 35.52 | 31.91 |
| MNMF | 73.45 | 53.15 | 47.16 | 47.15 | 39.40 | 69.44 | 60.99 | 58.28 | 58.90 | 54.44 | 71.55 | 61.17 | 62.50 | 54.51 | 56.62 |
| MCF | 74.62 | 54.61 | 53.24 | 50.98 | 39.92 | 69.36 | 62.88 | 56.03 | 53.08 | 52.30 | 79.46 | 61.80 | 57.96 | 56.91 | 52.82 |
| MLNMF | 77.27 | 64.89 | 63.74 | 55.15 | 41.59 | 65.72 | 60.61 | 54.84 | 50.70 | 50.49 | 68.95 | 56.43 | 58.20 | 56.80 | **57.94** |
| GMCF | 79.19 | 66.21 | 57.79 | 52.95 | 51.83 | 94.62 | 83.21 | 59.90 | 71.07 | 63.55 | 72.97 | 68.27 | 70.69 | 52.04 | 52.46 |
| Ours | **93.79** | **75.92** | **68.14** | **65.13** | **56.75** | **96.84** | **86.36** | **75.33** | **77.65** | **70.33** | **89.46** | 71.06 | **70.73** | **58.82** | 54.98 |
| Methods | F-score values | | | | | | | | | | | | | | |
| | ETH80 object database | | | | | COIL100 object database | | | | | MIT CBCL face database | | | | |
| | K=2 | K=3 | K=4 | K=5 | K=6 | K=2 | K=3 | K=4 | K=5 | K=6 | K=2 | K=3 | K=4 | K=5 | K=6 |
| GNMF | 86.27 | 61.50 | 53.54 | 43.41 | 32.00 | 82.72 | 81.59 | 62.41 | 64.73 | 54.64 | 71.17 | 42.61 | 39.78 | 32.84 | 28.69 |
| DNMF | 84.55 | 62.21 | 54.18 | 42.67 | 32.30 | 82.32 | 81.39 | 63.38 | 64.29 | 56.37 | 71.01 | 43.25 | 38.70 | 33.10 | 28.20 |
| LCCF | 91.15 | 58.55 | 60.88 | 58.04 | 53.28 | 68.35 | 65.76 | 56.36 | 66.38 | 62.67 | 69.89 | 67.53 | 59.14 | 51.53 | 53.51 |
| LCF | 91.49 | 56.48 | 57.49 | 38.89 | 27.23 | 89.11 | 76.29 | 63.98 | 66.57 | 59.81 | 58.29 | 38.18 | 37.22 | 33.61 | 27.14 |
| MNMF | 65.14 | 45.83 | 43.96 | 37.77 | 32.59 | 64.52 | 54.35 | 50.36 | 51.22 | 47.52 | 66.05 | 53.02 | 52.96 | 48.25 | 50.14 |
| MCF | 65.63 | 49.03 | 45.41 | 39.68 | 30.30 | 66.64 | 54.65 | 51.33 | 49.50 | 42.65 | 73.96 | 52.68 | 48.13 | 50.10 | 46.16 |
| MLNMF | 66.54 | 56.04 | 55.27 | 48.98 | 32.24 | 61.24 | 52.27 | 46.08 | 43.38 | 41.76 | 61.91 | 47.44 | 48.99 | 48.40 | 49.35 |
| GMCF | 75.62 | 62.52 | 60.32 | 50.81 | 46.41 | **94.75** | 81.49 | 64.39 | 69.79 | 63.20 | 72.11 | **68.41** | **66.69** | 43.99 | 55.65 |
| Ours | **93.13** | **69.58** | **63.73** | **58.80** | **53.59** | 92.89 | **82.87** | **69.45** | **75.90** | **66.12** | **87.59** | 66.25 | 57.40 | **50.86** | **56.21** |

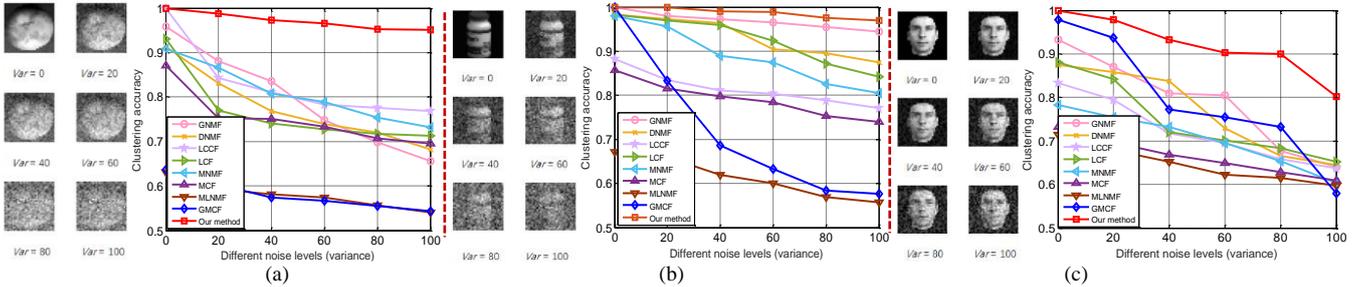

Figure 4: Clustering accuracies of each method on (a) ETH80, (b) COIL100, and (c) MIT CBCL against different levels of noises.

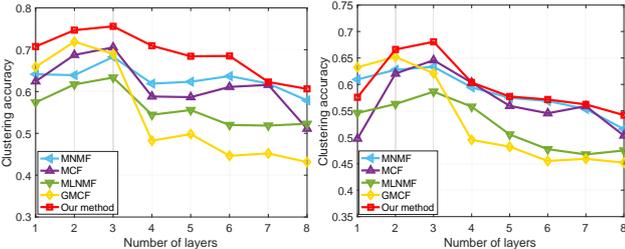

Figure 5: AC values vs. Number of layers on (left) ETH80 database and (right) MIT CBCL database.

ETH80, COIL100 and MIT CBCL are evaluated, which contains 3280/7200/3240 images from 80/100/10 classes respectively. For each evaluated method, we perform K-means clustering on the learnt new representation $V$. Specifically, following the common procedures [19-20], for each fixed K of clusters, we randomly choose K categories to form the input data matrix $X$ for representation learning and data clustering. In this study, the value of K varies from {2, 3, 4, 5, 6}. For each method, the rank r is set to K+1 as [20], and we average the numerical results over 30 random selections of K categories. The clustering results of AC and F-score over different values of K are shown in Table I. We find that: 1) the clustering results of each method usually decrease with the increasing of K; 2) deep matrix factorization methods (MNMF, MCF, MLNMF and GMCF) generally obtain the enhanced clustering performance than the single-layer GNMF, DNMF, LCCF and LCF for K=4/5/6, and our DSCF-Net delivers higher values of AC and F-score than other compared methods in most cases.

### C. Image Clustering against Corruptions

In this study, we prepare experiments to evaluate each method for clustering corrupted image data. To corrupt the data matrix $X$, we add random Gaussian noise with the variance being 0-100% with interval 20% into the selected pixels of images. Note that the position of the corrupted pixels it is unknown to users and the clustering accuracies by K-means clustering over corrupted noisy images with various levels of noise are illustrated in Fig.4. The results are obtained by choosing two categories and the AC values are averaged over 30 runs to avoid the randomness. We conclude that: 1) the AC values of each algorithm go down with the increasing noise levels in general; 2) our method outperforms the other algorithms in this study, which may be attributed to the fact that our DSCF-Net incorporates the error correction procedure into the representation learning and the factorization procedures are performed in the recovered clean data in each iteration.

### D. Clustering with Different Numbers of Layers

We investigate the effects of different number of layers on the clustering results of each multilayer model, i.e., MNMF, MCF, MLNMF, GMCF and our DSCF-Net. In this study, we aim to fix the data matrix, vary the number of layers from {1, 2, …, 8} and report the averaged clustering accuracies. ETH80 and MIT CBCL databases are used as the examples. For each database, we randomly choose four classes and average the results over 100 random selections to avoid the randomness. The AC values under different number of layers are shown in Fig.5, from which we can find that: 1) our method delivers the highest AC values in most cases; 2) generally, the AC value firstly increases with the increase of number of layers, but it start to decrease when the number of layers gets larger. It can also be found that most compared methods obtain the highest accuracy when the number of layers is 3, so we set the number of layers to 3 for all deep models for fair comparison.

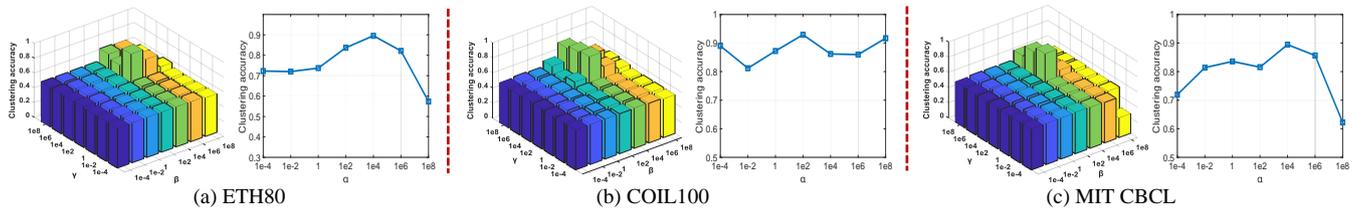

*Figure 6: Clustering accuracies vs. varied parameters of our DSCF-Net method based on the evaluated databases.*

*E. Parameter Sensitivity Analysis*

In this study, we explore the effects of model parameters on the clustering performance of DSCF-Net that has three trade-off parameters, i.e., $\alpha$, $\beta$ and $\gamma$. Since the optimal parameter selection is still an open issue, we adopt the widely-used grid search strategy [34-35] in our experiments to select the most important parameters. In this study, K is simply set to two, we randomly choose 2 categories to train our method. The results are averaged based on 30 random selections of categories and the central points in K-means clustering. The parameter selection results on MIT CBCL, ETH80 and COIL100 are illustrated in Fig.6, respectively. From the results, we find that the best clustering records are obtained based on similar parameter combinations, which is a good phenomenon of the model parameter selection. Finally, $\alpha=\beta=\gamma=10^4$ are used for MIT CBCL and ETH80, and $\alpha=10^2$, $\beta=\gamma=10^4$ are used for the COIL100 database in our experiments.

## VII. CONCLUDING REMARKS

We proposed a novel deep self-representative concept factorization network for unsupervised representation learning and clustering. DSCF-Net improved the representation and clustering abilities of deep factorization in threefold. First, to mine the hidden deep information, it employs a hierarchical factorization structure using multiple layers of linear transformations, where the hierarchical representation is formulated by optimizing the basis vectors in each layer to improve the representations indirectly. Second, to improve the robustness against noise, subspace recovery is integrated into the deep structures to recover the underlying visual subspace in which the basis concepts and representation are jointly optimized. Third, to obtain local representation, it uses an adaptive self-representative weighting strategy to preserve the locality of representation and avoid the tricky issue of neighbor selection. We evaluated DSCF-Net for image representation and clustering, and compared the results with related single-layer and multilayer models. Extensive results versified the effectiveness of DSCF-Net for representing and clustering images. In future, we will explore the strategy of updating the basis vectors and new representation in each layer jointly so that the representation will be more accurate.

ACKNOWLEDGMENTS

This work is partially supported by National Natural Science Foundation of China (61672365, 61871444 and 61806035), the Fundamental Research Funds for the Central Universities of China (JZ2019HGPA0102), and the Project Funded by the Priority Academic Program Development of Jiangsu Higher Education Institutions. Zhao Zhang is corresponding author.